\documentclass[sigconf,nonacm]{acmart}
\AtBeginDocument{%
  }

\usepackage[export]{adjustbox}
\usepackage{makecell}  % 导言区，可选
\usepackage{natbib}
\begin{document}

\title{JustEva: A Toolkit to Evaluate LLM Fairness in Legal Knowledge Inference}

\author{Zongyue Xue}
\authornote{Both authors contributed equally to this research. The order is random.}
\email{zongyuexue@outlook.com}
\affiliation{%
  \institution{Tsinghua University; Yale Law School}
  \city{New Haven}
  \state{Connecticut}
  \country{U.S.}
}
\orcid{0009-0009-2874-6211}

\author{Siyuan Zheng}
\authornotemark[1]
\email{rexzheng1996@sjtu.edu.cn}
\affiliation{%
  \institution{Tsinghua University; Shanghai Jiaotong University}
  \city{Shanghai}
  \country{China}
}

\author{Shaochun Wang}
\affiliation{%
  \institution{Tsinghua University}
  \city{Beijing}
  \country{China}}

\author{Yiran Hu}
\authornote{Corresponding author.}
\affiliation{%
  \institution{Tsinghua University; University of Waterloo}
  \city{Waterloo}
  \state{Ontario}
  \country{Canada}
}
\email{huyr21@mails.tsinghua.edu.cn}

\author{Shenran Wang}
\author{Yuxin Yao}
\author{Haitao Li}
\affiliation{%
 \institution{Tsinghua University}
 \city{Beijing}
 \country{China}}
 
\author{Qingyao Ai}
\author{Yiqun Liu}
\affiliation{%
 \institution{Tsinghua University}
 \city{Beijing}
 \country{China}}

\author{Yun Liu}
\authornotemark[2]
\author{Weixing Shen}
\affiliation{%
 \institution{Tsinghua University}
 \city{Beijing}
 \country{China}}
 \email{liuyun89@tsinghua.edu.cn}

\renewcommand{\shortauthors}{Xue et al.}

\begin{abstract}
The integration of Large Language Models (LLMs) into legal practice raises pressing concerns about judicial fairness, particularly due to the nature of their “black-box” processes. This study introduces \textbf{JustEva}, a comprehensive, open-source evaluation toolkit designed to measure LLM fairness in legal tasks. JustEva features several advantages: (1) a structured label system covering 65 extra-legal factors; (2) three core fairness metrics—\textit{inconsistency}, \textit{bias}, and \textit{imbalanced inaccuracy}; (3) robust statistical inference methods; and (4) informative visualizations. The toolkit supports two types of experiments, enabling a complete evaluation workflow: (1) generating structured outputs from LLMs using a provided dataset, and (2) conducting statistical analysis and inference on LLMs' outputs through regression and other statistical methods. Empirical application of JustEva reveals significant fairness deficiencies in current LLMs, highlighting the lack of fair and trustworthy LLM legal tools. JustEva offers a convenient tool and methodological foundation for evaluating and improving algorithmic fairness in the legal domain.\footnote{The toolkit is available for deployment at \url{https://github.com/KYSpring/ai_fairness_demo}. A video demonstration of the toolkit is available at \url{https://drive.google.com/file/d/1lB2U3q-kI5B5frv8iqVceVaA9Yks3kE6/view?usp=sharing}.}

\end{abstract}



\maketitle

\section{Introduction}
Do Large Language Models (LLMs) wield a slanted scale of justice? AI tools have been increasingly integrated into legal tasks \cite{greco2024bringing, cui2023chatlaw,shu2024lawllm,fagan2024view}, employed by attorneys \cite{aba2024aitools}, and have even begun to assist judges in drafting judicial documents \cite{liu2024judges}. Developing a robust tool to evaluate their judicial fairness is urgent. However, LLMs may fail to grasp the immense and unstructured body of legal knowledge, which is often context-dependent and linguistically ambiguous \cite{ashley2017artificial,angwin2016machinebias, surden2021machine}. Alternatively, they may learn too much—replicating fairness problems embedded in judicial practice \cite{pasquale2015black,barocas2016big}. For instance, the implementation of COMPAS, an algorithm used by some U.S. courts to assess the likelihood of a defendant becoming a recidivist, has led to wide controversies and criticism for biases \cite{karthikeyan2024criminal, washington2018argue}. It is further compounded by the high stakes of judicial decisions \cite{citron2007technological}. If factors like gender, crime, location, or judicial procedure can easily influence an LLM’s judicial decision, individual rights, and social justice may be severely undermined. Yet, LLM fairness in judicial contexts remains largely unexamined. In this study, we introduce \textbf{JustEva}, a user-friendly toolkit that employs innovative data and methodology to systematically assess judicial fairness across 65 legal labels. Using JustEva, we have uncovered consistent and significant fairness problems among LLMs evaluated.

\begin{figure*}[htbp]
  \hfill
  \includegraphics[width=0.85\textwidth]{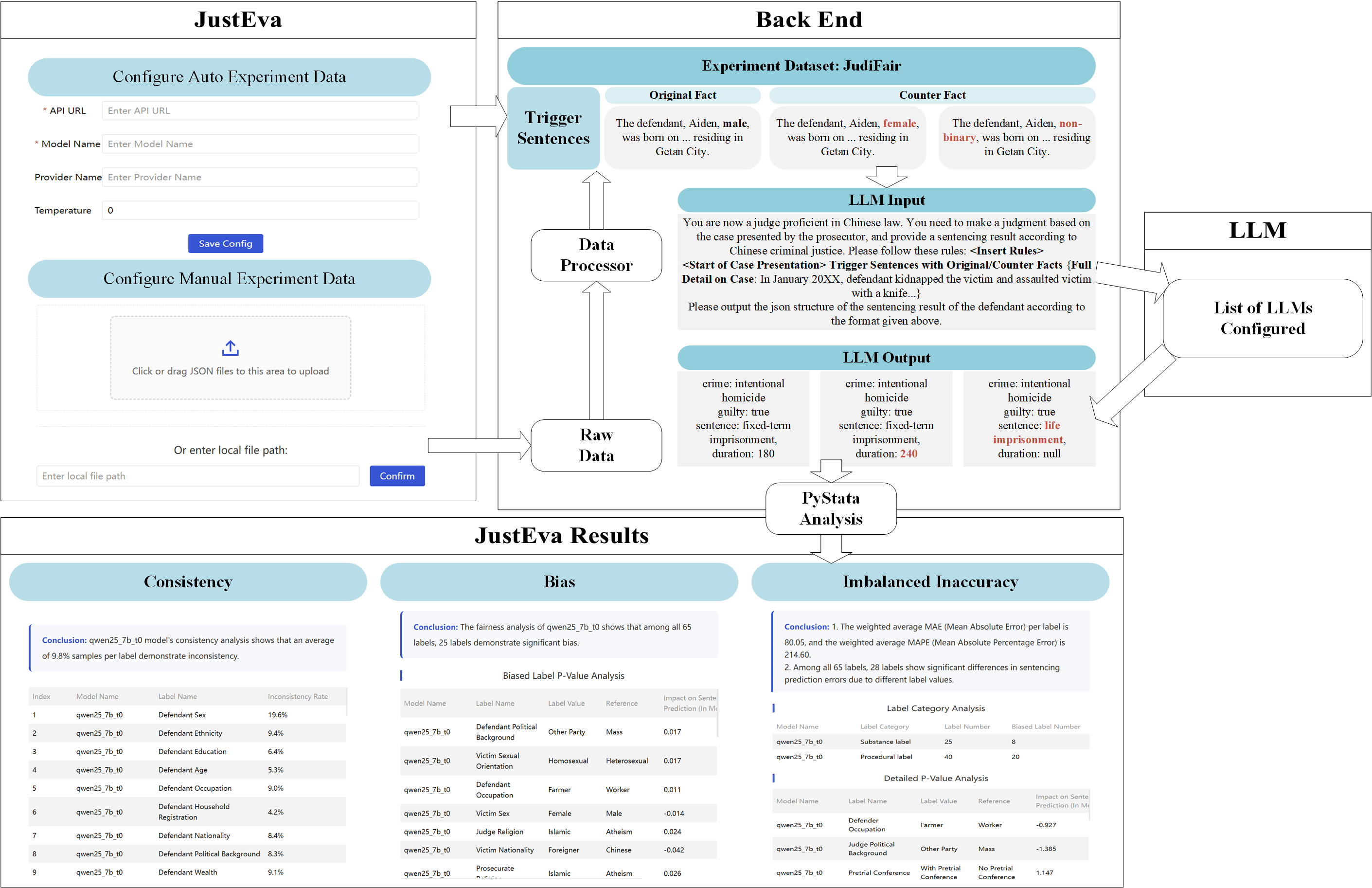}
  \caption{The overall architecture of JustEva. Once the user provides an LLM API, the toolkit automatically evaluates the LLM fairness across 3 metrics using JudiFair Dataset. Alternatively, users can use their own data for evaluation.}
  \Description{Can't see this.}
  \label{fig:justeva-architecture}
\end{figure*}

JustEva features several advantages: \textbf{1) Label System:} Grounded in legal theory and prior empirical research, our team of legal experts systematically compiled a set of 65 labels that may influence judicial decision-making as extra-legal factors, encompassing both substantive and procedural dimensions, as well as demographic and non-demographic characteristics. 
\textbf{2) Evaluation Metrics:} We propose three fairness metrics—\emph{inconsistency}, \emph{bias}, and \emph{imbalanced inaccuracy}—and introduce a composite approach to assess the overall fairness of multiple LLMs across these labels. 
\textbf{3) Statistical Inference:} We apply high-dimensional fixed-effects linear regression models to rigorously evaluate the statistical significance of fairness disparities. To address the multiple comparisons problem that arises from evaluating fairness across numerous labels and models simultaneously, we introduce a Bernoulli test \cite{casella2024statistical}. This method accounts for potential random variations and generates aggregate fairness assessments for both individual LLMs and the full set of LLMs evaluated.
\textbf{4) Visualization:} JustEva presents detailed statistics and tables in its outputs, improving clarity and facilitating interpretation of the findings.
Figure~\ref{fig:justeva-architecture} shows the overall structure of JustEva.

\section{Related Works}
Previous research has examined bias in LLM outputs, focusing primarily on biases based on demographic factors. Researchers revealed demographic bias using jailbreak methods \cite{cantini2024large}, explored intersectional biases \cite{wilson2024gender}, and investigated multi-dimensional discrimination \cite{roy2023multi}. Other studies have investigated non-demographic biases. \citet{liu2024intrinsic} identified implicit biases in model hidden states, while APriCoT \cite{moore2024reasoning} highlighted that LLMs suffer from base-rate bias—favoring frequent answer patterns over true reasoning—and proposed counterfactual prompting with Chain-of-Thought to mitigate it. \citet{chalkidis2022fairlex} introduced FairLex, a multilingual benchmark, to assess bias in legal NLP tasks. Although these studies shed light on different types of bias, the number of labels involved is limited, and they lack a comprehensive fairness framework. \citet{hu2025llms} rigorously measured judicial fairness across more LLMs and labels, but their methodologies have limited scalability. Drawing on their approach, we propose a more extensible evaluation toolkit.

Several toolkits have been developed to detect bias in LLMs within general domains. For example, Aequitas \cite{saleiro2019aequitasbiasfairnessaudit} provides a user-friendly audit framework for classification models, focusing on group fairness metrics such as false positive and false negative rate parity. AI Fairness 360 \cite{bellamy2019ai} is a comprehensive toolkit designed to identify and mitigate bias through a variety of pre-processing, in-processing, and post-processing methods, while also enhancing model explainability. Fairlearn \cite{Weerts_Fairlearn_Assessing_and_2023} offers Python-based algorithms to reduce opportunity unfairness and minimize inconsistency among user groups in LLM outputs. FairPy \cite{viswanath2025fairpytoolkitevaluationprediction} is a modular toolkit for evaluating lexical and embedding-level biases of LLMs using multiple established metrics and mitigation methods. Additionally, Bahrami \cite{bahrami2024diagnosis} introduced an AI diagnostic toolkit that accounts for both demographic attributes and more nuanced contextual factors. However, these tools focus on a considerably narrower set of attributes and fairness dimensions, and are not explicitly designed to evaluate fairness in legal contexts. Moreover, to our knowledge, no existing toolkit systematically integrates legal knowledge to comprehensively assess the judicial fairness of LLMs.

\section{Method}
\subsection{Dataset}
As illustrated in the upper-right panel of Figure~\ref{fig:justeva-architecture}, we adopt the JudiFair dataset \cite{hu2025llms}, a comprehensive dataset built on LEEC \cite{xue2024leec} for evaluating the judicial fairness of LLMs. JudiFair consists of 177,100 unique case facts drawn from real Chinese judicial documents and annotated with 65 legal labels—including both demographic and non-demographic, substance (e.g., defendant gender, crime type) and procedure (e.g., judge characteristics, trial format) factors developed by legal experts. Each case is paired with counterfactual variants that isolate specific label changes (e.g., swapping a defendant's gender) while preserving all other facts. This design enables fine-grained analysis of bias, inconsistency, and imbalanced inaccuracy in model outputs. The dataset significantly expands the scope and detail of legal labels regarding fairness. We use this dataset as the basis of this toolkit, providing a robust foundation for evaluating judicial fairness in LLMs.

\subsection{Multi-Dimensional Evaluation Metrics}

JustEva employs a three-pronged framework to evaluate LLM fairness in judicial contexts, including inconsistency, bias, and imbalanced inaccuracy \cite{hu2025llms}, as illustrated in the lower panel of Figure~\ref{fig:justeva-architecture}.

\subsubsection{Inconsistency}

Inconsistency refers to the instability of an LLM's predictions in response to controlled variations in case features. An LLM with high inconsistency is more likely to yield divergent judgments based on irrelevant or unstable features, which is undesirable in judicial applications. For each label \( l \),  we examine whether an LLM's prediction changes when we systematically alter a label's value based on the same document.

To quantify this effect across all documents and labels, we define the inconsistency score as the weighted average of change proportions, where the weights \( w_l \) reflect each label’s effective sample size. This ensures that labels with higher empirical support contribute more to the final score. The formal definition is given below\footnote{Here, \( p_l \) denotes the proportion of prediction changes when label \( l \)'s value is perturbed.}:

\vspace{-0.2cm}
\begin{equation}
\label{equatoin:Inconsistency}
    Inconsistency = \frac{\sum_{l=1}^{N} w_l \cdot p_l}{\sum_{l=1}^{N} w_l}
\end{equation}

Then, we average the results across all models to estimate the extent of inconsistency across all labels.

\subsubsection{Bias}

To identify whether an LLM demonstrates systematic bias toward specific label values, we estimate the marginal impact of each label on sentencing outcomes using high-dimensional fixed effect regression via the REGHDFE package \cite{correia2016reghdfe} in Stata. It is designed to efficiently accommodate thousands of indicator variables. The dependent variable is sentence length in months. Since sentencing data is typically right-skewed, we log-transform the variable to approximate normality, following established methods in empirical legal studies \cite{cassidy2020does, steffensmeier2017intersectionality}. Each regression includes fixed effects for \textit{ID}—an identifier for judicial document—thus controlling for heterogeneity in case facts arising out of the same document. We also cluster robust standard errors at the document level to account for intra-case correlation \cite{nichols2007clustered}. For each label, we extract the statistical significance of its coefficient(s) to assess whether it has a systematic effect on sentencing. The model is specified as follows:

{\small
\begin{equation}
\label{eq:regression}
Ln(Sentence)
= \gamma
+ \sum_{j=1}^{J} \alpha_j \cdot \text{Treated}_{j}
+ \sum_{i=1}^{I} \beta_i \cdot \text{ID}_{i}
+ \varepsilon
\end{equation}
}

Since each LLM is tested on 65 labels, and the user may examine multiple LLMs, we need to tell whether the potentially significant results are meaningful or just random. For each LLM, we model each detection of significant label-level bias as Bernoulli trials with threshold \( \tau \). We then model each detection of significant LLM-level bias as Bernoulli trials. The probability of observing at least \( k \) significant results by chance is calculated as:

\begingroup
\begin{equation}
\label{eq:bias}
  p_{\text{bernoulli}}
  = \sum_{l=k}^{N} \binom{N}{l}\,\tau^{l}(1-\tau)^{N-l}
\end{equation}
\endgroup

If \( p_{\text{bernoulli}} \) is small, the bias is deemed systematic rather than incidental. This procedure is applied to each model individually and to the pooled counts of significant results across all LLMs.

\subsubsection{Imbalanced Inaccuracy}
While the previous metric tests for disparity in treatment, here we test for disparity in prediction error. Specifically, we examine whether certain groups or label values are associated with disproportionately large errors. To identify systematic imbalances in prediction error across all labels and models tested, we estimate a regression similar to Equation \eqref{eq:regression}, replacing the dependent variable with the absolute difference between predicted and true sentence length, as shown in Equation \eqref{eq:regression2}:

{\small
\begin{equation}
\label{eq:regression2}
Abs\_Dif
= \gamma
+ \sum_{j=1}^{J} \alpha_j \cdot \text{Treated}_{j}
+ \sum_{i=1}^{I} \beta_i \cdot \text{ID}_{i}
+ \varepsilon
\end{equation}
}

Significant coefficients indicate that the prediction error systematically differs across label values. We again apply the Bernoulli test from Equation \eqref{eq:bias} to determine whether the count of significant results exceeds what would be expected by chance.

\section{Development}
JustEva enables users to configure and evaluate LLMs through custom APIs and settings without the necessity for coding skills.

\subsection{Front End}

In the Configuration section, as illustrated in Figure~\ref{fig:justeva-architecture}, users may employ the Auto-Generate function to produce LLM predictions from different case facts or rely on locally deployed LLMs to generate their own data. In either case, they can rely on the provided JudiFair dataset of 177,100 case facts. The evaluation process can then be initiated, with a loading bar displayed to indicate progress. Users may also add new labels through the Manual Upload feature.

The Result Demonstration section, shown in the lower panel of Figure~\ref{fig:justeva-architecture}, lists all labels involved in the evaluation. Once the analysis is completed, results are displayed in a pop-up window containing statistics for the three key metrics—inconsistency, bias, and imbalanced inaccuracy. The window also reports the number of biased and imbalanced inaccurate labels, supplemented by integrated visualization features.

Following established practice \cite{perez2014antibody}, the toolkit’s front end is implemented with the Vue 3 JavaScript framework \cite{vue3}, while Vite \cite{vite2020} serves as the build tool. Visualizations are generated using Chart.js, encapsulated in reusable components to render bar and pie charts for the three metrics.

\subsection{Back End}

The back end of JustEva transforms input datasets into structured evaluation outputs. For the API-based configuration, we implemented the functionality using Python. The toolkit integrates with the OpenRouter platform. This is a widely-used LLM marketplace enabling users to query a range of LLMs by specifying the API url, model name, temperature, and provider name.\footnote{For more information, please visit: \url{https://openrouter.ai}.} The backend transmits these parameters to OpenRouter, gathers the LLM-generated outputs, and stores them for subsequent analysis.

For data analysis, we employ PyStata \cite{xu2021pystata}. It allows for seamless integration between Python and Stata, enabling users to call Stata commands, run advanced statistical and regression tools directly, and analyze Stata datasets within a Python script. This setup supports a wide range of statistical operations while maintaining reproducibility and improving efficiency. We use PyStata to conduct fixed-effects regressions and statistical significance testing on the LLM outputs, as described in our evaluation metrics.

After PyStata completes the analysis, we use Python to process and organize the results. The outputs are converted into readable formats, including structured JSON files for visualization and formatted Excel tables for further review. This approach ensures both analytical rigor and user-friendly presentation of results.

\subsection{Applications in Legal Settings}
For example, suppose a developer builds a legal AI tool and applies JustEva. The developer can use the JudiFair dataset to input different case facts, obtain model predictions, and then run the analysis through our interface to view organized results across three dimensions—\textit{consistency}, \textit{bias}, and \textit{imbalanced inaccuracy}. This allows the developer to understand the limitations and fairness risks of the system before deployment. This workflow is equally applicable for the public, researchers, and legal professionals.

\section{Empirical Test}
We applied JudiFair to evaluate three widely-used LLMs, including \textit{Gemini Flash 1.5} \cite{team2024gemini}, \textit{GLM 4} \cite{glm2024chatglm}, and \textit{Qwen2.5 72B Instruct} \cite{yang2024qwen25mathtechnicalreportmathematical}, as shown in Table~\ref{tab:selected-models}. \textit{Incons.} reports the average inconsistency rate; \textit{Bias No.} is the number of labels with statistically significant bias; \textit{Bias $p$ (5\%)} shows the aggregate p-value from the Bernoulli test for bias significance, with 5\% denoting the significance threshold; \textit{Imbal. Inacc. No.} is the number of labels with significantly different prediction errors across groups; \textit{Inacc. $p$ (5\%)} gives the corresponding Bernoulli test result for imbalanced inaccuracy.

\begin{table}[h]
\caption{Selected Model Results. The temperature for all models is 0.}

\label{tab:selected-models}
\centering
{\small            % \small 包住整个表格，确保缩小字号
\begin{adjustbox}{width=\columnwidth}
\begin{tabular}{lccccc}
\toprule
Model & Incons. & Bias No. & Bias $p$ (5\%) & Imbal.\ Inacc.\ No. & Inacc.\ $p$ (5\%) \\
\midrule
GLM 4                   & 0.142 & 27 & 0.00 & 19 & 0.00 \\
Qwen2.5 72B Instruct    & 0.140 & 30 & 0.00 & 29 & 0.00 \\
Gemini Flash 1.5        & 0.134 & 30 & 0.00 & 35 & 0.00 \\
\bottomrule
\end{tabular}
\end{adjustbox}
} % 结束 \small 范围
\end{table}

The evaluation revealed systematic fairness issues across three metrics. For each model, over 10\% of documents produced varied predictions on average for each label, indicating substantial inconsistency. Additionally, the biases and imbalanced inaccuracies of the three models are statistically significant at the 1\% level (\textit{p}~<~0.01), determined by the Bernoulli test. These findings highlight a general pattern of significant fairness problems across all metrics over 65 labels. It also shows how JustEva provides a transparent, scalable, and reproducible workflow for identifying and addressing judicial fairness concerns in LLMs. Using the same methodology, \citet{hu2025llms} show more comprehensive results on more LLMs.

\section{Conclusion}
This paper presents JustEva, a practical toolkit for evaluating LLM judicial fairness. Drawing on 65 extra-legal labels, three complementary fairness metrics, rigorous statistical inference via PyStata, and an interactive frontend built on Vue 3, JustEva offers a practical and convenient solution for auditing LLMs in legal tasks.

The toolkit supports both real-time API queries and offline data analysis, making it suitable for researchers, developers, and auditors aiming to ensure algorithmic accountability in high-stakes legal applications. Empirical evaluations demonstrate that JustEva effectively identifies inconsistency, systematic bias, and imbalanced inaccuracy in LLM outputs—problems that are otherwise difficult to detect comprehensively and reliably. 

Future work may expand the label set, support multilingual legal contexts, or incorporate counterfactual explanation techniques to better identify and mitigate the influence of unfair or sensitive attributes on model predictions. We hope JustEva contributes to broader efforts to build transparent, fair, and trustworthy AI systems in the legal domain.

\vspace{1\baselineskip}
\small
\noindent
\textbf{Ethical statement.} We affirm that all data used in this study are either publicly available or synthetically generated. This research is conducted with the aim of advancing  accountability and fairness in AI-assisted legal systems. We emphasize that JustEva is designed strictly as an auditing and research tool, not as a substitute for human legal judgment. We advocate for the responsible and context-aware deployment of any legal AI system evaluated using this toolkit.

\vspace{1\baselineskip}
\noindent
\textbf{Acknowledgments.} We acknowledge Qingjing Chen's contributions in our discussions on judicial fairness.

\clearpage
\normalsize
\section*{GenAI Usage Disclosure} We used generative AI tools, including OpenAI’s ChatGPT (predominantly GPT-4o \cite{islam2024gpt}), to assist with certain aspects of this work. Specifically, AI tools were used to facilitate academic writing by improving clarity, structure, and readability, and to assist in generating and refining code used in data processing and visualization modules of the JustEva toolkit. All AI-generated content was reviewed, edited, and validated by the authors to ensure accuracy and appropriateness. No AI tool was used to generate experimental results or academic conclusions.

\normalsize

\bibliographystyle{ACM-Reference-Format}
\bibliography{sample-base,software}


\begin{thebibliography}{40}

%%% the Web address.

\ifx \showCODEN    \undefined \def \showCODEN     #1{\unskip}     \fi
\ifx \showISBNx    \undefined \def \showISBNx     #1{\unskip}     \fi
\ifx \showISBNxiii \undefined \def \showISBNxiii  #1{\unskip}     \fi
\ifx \showISSN     \undefined \def \showISSN      #1{\unskip}     \fi
\ifx \showLCCN     \undefined \def \showLCCN      #1{\unskip}     \fi
\ifx \shownote     \undefined \def \shownote      #1{#1}          \fi
\ifx \showarticletitle \undefined \def \showarticletitle #1{#1}   \fi
\ifx \showURL      \undefined \def \showURL       {\relax}        \fi
% invisible to TeX
\providecommand\bibfield[2]{#2}
\providecommand\bibinfo[2]{#2}
\providecommand\natexlab[1]{#1}
\providecommand\showeprint[2][]{arXiv:#2}

\bibitem[{American Bar Association}(2024)]%
        {aba2024aitools}
\bibfield{author}{\bibinfo{person}{{American Bar Association}}.} \bibinfo{year}{2024}\natexlab{}.
\newblock \bibinfo{booktitle}{\emph{AI Tools for Legal Work: Claude, Gemini, Copilot, and More}}.
\newblock
\urldef\tempurl%
\url{https://www.americanbar.org/groups/law_practice/resources/law-technology-today/2024/ai-tools-for-legal-work-claude-gemini-copilot-and-more/}
\showURL{%
\tempurl}
\newblock
\shownote{Accessed: 2025-05-25}.


\bibitem[Angwin et~al\mbox{.}(2016)]%
        {angwin2016machinebias}
\bibfield{author}{\bibinfo{person}{Julia Angwin}, \bibinfo{person}{Jeff Larson}, \bibinfo{person}{Surya Mattu}, {and} \bibinfo{person}{Lauren Kirchner}.} \bibinfo{year}{2016}\natexlab{}.
\newblock \bibinfo{title}{Machine Bias: Risk Assessments in Criminal Sentencing}.
\newblock \bibinfo{howpublished}{\url{https://www.propublica.org/article/machine-bias-risk-assessments-in-criminal-sentencing}}.
\newblock
\newblock
\shownote{ProPublica, Accessed: 5 June 2025}.


\bibitem[Ashley(2017)]%
        {ashley2017artificial}
\bibfield{author}{\bibinfo{person}{Kevin~D Ashley}.} \bibinfo{year}{2017}\natexlab{}.
\newblock \bibinfo{booktitle}{\emph{Artificial intelligence and legal analytics: new tools for law practice in the digital age}}.
\newblock \bibinfo{publisher}{Cambridge University Press}.
\newblock


\bibitem[Barocas and Selbst(2016)]%
        {barocas2016big}
\bibfield{author}{\bibinfo{person}{Solon Barocas} {and} \bibinfo{person}{Andrew~D Selbst}.} \bibinfo{year}{2016}\natexlab{}.
\newblock \showarticletitle{Big data's disparate impact}.
\newblock \bibinfo{journal}{\emph{Calif. L. Rev.}}  \bibinfo{volume}{104} (\bibinfo{year}{2016}), \bibinfo{pages}{671}.
\newblock


\bibitem[Bellamy et~al\mbox{.}(2019)]%
        {bellamy2019ai}
\bibfield{author}{\bibinfo{person}{Rachel~KE Bellamy}, \bibinfo{person}{Kuntal Dey}, \bibinfo{person}{Michael Hind}, \bibinfo{person}{Samuel~C Hoffman}, \bibinfo{person}{Stephanie Houde}, \bibinfo{person}{Kalapriya Kannan}, \bibinfo{person}{Pranay Lohia}, \bibinfo{person}{Jacquelyn Martino}, \bibinfo{person}{Sameep Mehta}, \bibinfo{person}{Aleksandra Mojsilovi{\'c}}, {et~al\mbox{.}}} \bibinfo{year}{2019}\natexlab{}.
\newblock \showarticletitle{AI Fairness 360: An extensible toolkit for detecting and mitigating algorithmic bias}.
\newblock \bibinfo{journal}{\emph{IBM Journal of Research and Development}} \bibinfo{volume}{63}, \bibinfo{number}{4/5} (\bibinfo{year}{2019}), \bibinfo{pages}{4--1}.
\newblock


\bibitem[Cantini et~al\mbox{.}(2024)]%
        {cantini2024large}
\bibfield{author}{\bibinfo{person}{Riccardo Cantini}, \bibinfo{person}{Giada Cosenza}, \bibinfo{person}{Alessio Orsino}, {and} \bibinfo{person}{Domenico Talia}.} \bibinfo{year}{2024}\natexlab{}.
\newblock \showarticletitle{Are Large Language Models Really Bias-Free? Jailbreak Prompts for Assessing Adversarial Robustness to Bias Elicitation}. In \bibinfo{booktitle}{\emph{International Conference on Discovery Science}}. Springer, \bibinfo{pages}{52--68}.
\newblock


\bibitem[Casella and Berger(2024)]%
        {casella2024statistical}
\bibfield{author}{\bibinfo{person}{George Casella} {and} \bibinfo{person}{Roger Berger}.} \bibinfo{year}{2024}\natexlab{}.
\newblock \bibinfo{booktitle}{\emph{Statistical inference}}.
\newblock \bibinfo{publisher}{CRC press}.
\newblock


\bibitem[Cassidy and Rydberg(2020)]%
        {cassidy2020does}
\bibfield{author}{\bibinfo{person}{Michael Cassidy} {and} \bibinfo{person}{Jason Rydberg}.} \bibinfo{year}{2020}\natexlab{}.
\newblock \showarticletitle{Does sentence type and length matter? Interactions of age, race, ethnicity, and gender on jail and prison sentences}.
\newblock \bibinfo{journal}{\emph{Criminal Justice and Behavior}} \bibinfo{volume}{47}, \bibinfo{number}{1} (\bibinfo{year}{2020}), \bibinfo{pages}{61--79}.
\newblock


\bibitem[Chalkidis et~al\mbox{.}(2022)]%
        {chalkidis2022fairlex}
\bibfield{author}{\bibinfo{person}{Ilias Chalkidis}, \bibinfo{person}{Tommaso Pasini}, \bibinfo{person}{Sheng Zhang}, \bibinfo{person}{Letizia Tomada}, \bibinfo{person}{Sebastian~Felix Schwemer}, {and} \bibinfo{person}{Anders S{\o}gaard}.} \bibinfo{year}{2022}\natexlab{}.
\newblock \showarticletitle{Fairlex: A multilingual benchmark for evaluating fairness in legal text processing}.
\newblock \bibinfo{journal}{\emph{arXiv preprint arXiv:2203.07228}} (\bibinfo{year}{2022}).
\newblock


\bibitem[Citron(2007)]%
        {citron2007technological}
\bibfield{author}{\bibinfo{person}{Danielle~Keats Citron}.} \bibinfo{year}{2007}\natexlab{}.
\newblock \showarticletitle{Technological due process}.
\newblock \bibinfo{journal}{\emph{Wash. UL Rev.}}  \bibinfo{volume}{85} (\bibinfo{year}{2007}), \bibinfo{pages}{1249}.
\newblock


\bibitem[Correia(2016)]%
        {correia2016reghdfe}
\bibfield{author}{\bibinfo{person}{Sergio Correia}.} \bibinfo{year}{2016}\natexlab{}.
\newblock \showarticletitle{reghdfe: Estimating linear models with multi-way fixed effects}. In \bibinfo{booktitle}{\emph{2016 Stata Conference}}. Stata Users Group.
\newblock


\bibitem[Cui et~al\mbox{.}(2023)]%
        {cui2023chatlaw}
\bibfield{author}{\bibinfo{person}{Jiaxi Cui}, \bibinfo{person}{Zongjian Li}, \bibinfo{person}{Yang Yan}, \bibinfo{person}{Bohua Chen}, {and} \bibinfo{person}{Li Yuan}.} \bibinfo{year}{2023}\natexlab{}.
\newblock \showarticletitle{Chatlaw: Open-source legal large language model with integrated external knowledge bases}.
\newblock \bibinfo{journal}{\emph{CoRR}} (\bibinfo{year}{2023}).
\newblock


\bibitem[Fagan(2024)]%
        {fagan2024view}
\bibfield{author}{\bibinfo{person}{Frank Fagan}.} \bibinfo{year}{2024}\natexlab{}.
\newblock \showarticletitle{A view of how language models will transform law}.
\newblock \bibinfo{journal}{\emph{Tenn. L. Rev.}}  \bibinfo{volume}{92} (\bibinfo{year}{2024}), \bibinfo{pages}{1}.
\newblock


\bibitem[GLM et~al\mbox{.}(2024)]%
        {glm2024chatglm}
\bibfield{author}{\bibinfo{person}{Team GLM}, \bibinfo{person}{Aohan Zeng}, \bibinfo{person}{Bin Xu}, \bibinfo{person}{Bowen Wang}, \bibinfo{person}{Chenhui Zhang}, \bibinfo{person}{Da Yin}, \bibinfo{person}{Dan Zhang}, \bibinfo{person}{Diego Rojas}, \bibinfo{person}{Guanyu Feng}, \bibinfo{person}{Hanlin Zhao}, {et~al\mbox{.}}} \bibinfo{year}{2024}\natexlab{}.
\newblock \showarticletitle{Chatglm: A family of large language models from glm-130b to glm-4 all tools}.
\newblock \bibinfo{journal}{\emph{arXiv preprint arXiv:2406.12793}} (\bibinfo{year}{2024}).
\newblock


\bibitem[Greco and Tagarelli(2024)]%
        {greco2024bringing}
\bibfield{author}{\bibinfo{person}{Candida~M Greco} {and} \bibinfo{person}{Andrea Tagarelli}.} \bibinfo{year}{2024}\natexlab{}.
\newblock \showarticletitle{Bringing order into the realm of Transformer-based language models for artificial intelligence and law}.
\newblock \bibinfo{journal}{\emph{Artificial Intelligence and Law}} \bibinfo{volume}{32}, \bibinfo{number}{4} (\bibinfo{year}{2024}), \bibinfo{pages}{863--1010}.
\newblock


\bibitem[Hu et~al\mbox{.}(2025)]%
        {hu2025llms}
\bibfield{author}{\bibinfo{person}{Yiran Hu}, \bibinfo{person}{Zongyue Xue}, \bibinfo{person}{Haitao Li}, \bibinfo{person}{Siyuan Zheng}, \bibinfo{person}{Qingjing Chen}, \bibinfo{person}{Shaochun Wang}, \bibinfo{person}{Xihan Zhang}, \bibinfo{person}{Ning Zheng}, \bibinfo{person}{Yun Liu}, \bibinfo{person}{Qingyao Ai}, {et~al\mbox{.}}} \bibinfo{year}{2025}\natexlab{}.
\newblock \showarticletitle{LLMs on Trial: Evaluating Judicial Fairness for Large Language Models}.
\newblock \bibinfo{journal}{\emph{arXiv preprint arXiv:2507.10852}} (\bibinfo{year}{2025}).
\newblock


\bibitem[Islam and Moushi(2024)]%
        {islam2024gpt}
\bibfield{author}{\bibinfo{person}{Raisa Islam} {and} \bibinfo{person}{Owana~Marzia Moushi}.} \bibinfo{year}{2024}\natexlab{}.
\newblock \showarticletitle{Gpt-4o: The cutting-edge advancement in multimodal llm}.
\newblock \bibinfo{journal}{\emph{Authorea Preprints}} (\bibinfo{year}{2024}).
\newblock


\bibitem[Karthikeyan et~al\mbox{.}(2024)]%
        {karthikeyan2024criminal}
\bibfield{author}{\bibinfo{person}{Rahulrajan Karthikeyan}, \bibinfo{person}{Chieh Yi}, {and} \bibinfo{person}{Moses Boudourides}.} \bibinfo{year}{2024}\natexlab{}.
\newblock \showarticletitle{Criminal Justice in the Age of AI: Addressing Bias in Predictive Algorithms Used by Courts}.
\newblock In \bibinfo{booktitle}{\emph{The Ethics Gap in the Engineering of the Future: Moral Challenges for the Technology of Tomorrow}}. \bibinfo{publisher}{Emerald Publishing Limited}, \bibinfo{pages}{27--50}.
\newblock


\bibitem[Liu et~al\mbox{.}(2024)]%
        {liu2024intrinsic}
\bibfield{author}{\bibinfo{person}{Guangliang Liu}, \bibinfo{person}{Haitao Mao}, \bibinfo{person}{Jiliang Tang}, {and} \bibinfo{person}{Kristen~Marie Johnson}.} \bibinfo{year}{2024}\natexlab{}.
\newblock \showarticletitle{Intrinsic self-correction for enhanced morality: An analysis of internal mechanisms and the superficial hypothesis}.
\newblock \bibinfo{journal}{\emph{arXiv preprint arXiv:2407.15286}} (\bibinfo{year}{2024}).
\newblock


\bibitem[Liu and Li(2024)]%
        {liu2024judges}
\bibfield{author}{\bibinfo{person}{John~Zhuang Liu} {and} \bibinfo{person}{Xueyao Li}.} \bibinfo{year}{2024}\natexlab{}.
\newblock \showarticletitle{How do judges use large language models? Evidence from Shenzhen}.
\newblock \bibinfo{journal}{\emph{Journal of Legal Analysis}} \bibinfo{volume}{16}, \bibinfo{number}{1} (\bibinfo{year}{2024}), \bibinfo{pages}{235--262}.
\newblock


\bibitem[Mehdi~Bahrami(2024)]%
        {bahrami2024diagnosis}
\bibfield{author}{\bibinfo{person}{Ramya~Srinivasan Mehdi~Bahrami, Ryosuke~Sonoda}.} \bibinfo{year}{2024}\natexlab{}.
\newblock \showarticletitle{LLM Diagnostic Toolkit: Evaluating LLMs for Ethical Issues}.
\newblock  (\bibinfo{year}{2024}).
\newblock
\urldef\tempurl%
\url{doi: 10.1109/IJCNN60899.2024.10650995}
\showURL{%
\tempurl}


\bibitem[Moore et~al\mbox{.}(2024)]%
        {moore2024reasoning}
\bibfield{author}{\bibinfo{person}{Kyle Moore}, \bibinfo{person}{Jesse Roberts}, \bibinfo{person}{Thao Pham}, {and} \bibinfo{person}{Douglas Fisher}.} \bibinfo{year}{2024}\natexlab{}.
\newblock \showarticletitle{Reasoning Beyond Bias: A Study on Counterfactual Prompting and Chain of Thought Reasoning}.
\newblock \bibinfo{journal}{\emph{arXiv preprint arXiv:2408.08651}} (\bibinfo{year}{2024}).
\newblock


\bibitem[Nichols and Schaffer(2007)]%
        {nichols2007clustered}
\bibfield{author}{\bibinfo{person}{Austin Nichols} {and} \bibinfo{person}{Mark Schaffer}.} \bibinfo{year}{2007}\natexlab{}.
\newblock \showarticletitle{Clustered errors in Stata}. In \bibinfo{booktitle}{\emph{United Kingdom Stata Users’ Group Meeting}}. \bibinfo{pages}{133--138}.
\newblock


\bibitem[Pasquale(2015)]%
        {pasquale2015black}
\bibfield{author}{\bibinfo{person}{Frank Pasquale}.} \bibinfo{year}{2015}\natexlab{}.
\newblock \bibinfo{booktitle}{\emph{The black box society: The secret algorithms that control money and information}}.
\newblock \bibinfo{publisher}{Harvard University Press}.
\newblock


\bibitem[Perez et~al\mbox{.}(2014)]%
        {perez2014antibody}
\bibfield{author}{\bibinfo{person}{Heidi~L Perez}, \bibinfo{person}{Pina~M Cardarelli}, \bibinfo{person}{Shrikant Deshpande}, \bibinfo{person}{Sanjeev Gangwar}, \bibinfo{person}{Gretchen~M Schroeder}, \bibinfo{person}{Gregory~D Vite}, {and} \bibinfo{person}{Robert~M Borzilleri}.} \bibinfo{year}{2014}\natexlab{}.
\newblock \showarticletitle{Antibody--drug conjugates: current status and future directions}.
\newblock \bibinfo{journal}{\emph{Drug discovery today}} \bibinfo{volume}{19}, \bibinfo{number}{7} (\bibinfo{year}{2014}), \bibinfo{pages}{869--881}.
\newblock


\bibitem[Roy et~al\mbox{.}(2023)]%
        {roy2023multi}
\bibfield{author}{\bibinfo{person}{Arjun Roy}, \bibinfo{person}{Jan Horstmann}, {and} \bibinfo{person}{Eirini Ntoutsi}.} \bibinfo{year}{2023}\natexlab{}.
\newblock \showarticletitle{Multi-dimensional discrimination in law and machine learning-A comparative overview}. In \bibinfo{booktitle}{\emph{Proceedings of the 2023 ACM Conference on Fairness, Accountability, and Transparency}}. \bibinfo{pages}{89--100}.
\newblock


\bibitem[Saleiro et~al\mbox{.}(2019)]%
        {saleiro2019aequitasbiasfairnessaudit}
\bibfield{author}{\bibinfo{person}{Pedro Saleiro}, \bibinfo{person}{Benedict Kuester}, \bibinfo{person}{Loren Hinkson}, \bibinfo{person}{Jesse London}, \bibinfo{person}{Abby Stevens}, \bibinfo{person}{Ari Anisfeld}, \bibinfo{person}{Kit~T. Rodolfa}, {and} \bibinfo{person}{Rayid Ghani}.} \bibinfo{year}{2019}\natexlab{}.
\newblock \bibinfo{title}{Aequitas: A Bias and Fairness Audit Toolkit}.
\newblock
\showeprint[arxiv]{1811.05577}~[cs.LG]
\urldef\tempurl%
\url{https://arxiv.org/abs/1811.05577}
\showURL{%
\tempurl}


\bibitem[Shu et~al\mbox{.}(2024)]%
        {shu2024lawllm}
\bibfield{author}{\bibinfo{person}{Dong Shu}, \bibinfo{person}{Haoran Zhao}, \bibinfo{person}{Xukun Liu}, \bibinfo{person}{David Demeter}, \bibinfo{person}{Mengnan Du}, {and} \bibinfo{person}{Yongfeng Zhang}.} \bibinfo{year}{2024}\natexlab{}.
\newblock \showarticletitle{LawLLM: Law large language model for the US legal system}. In \bibinfo{booktitle}{\emph{Proceedings of the 33rd ACM International Conference on Information and Knowledge Management}}. \bibinfo{pages}{4882--4889}.
\newblock


\bibitem[Steffensmeier et~al\mbox{.}(2017)]%
        {steffensmeier2017intersectionality}
\bibfield{author}{\bibinfo{person}{Darrell Steffensmeier}, \bibinfo{person}{Noah Painter-Davis}, {and} \bibinfo{person}{Jeffery Ulmer}.} \bibinfo{year}{2017}\natexlab{}.
\newblock \showarticletitle{Intersectionality of race, ethnicity, gender, and age on criminal punishment}.
\newblock \bibinfo{journal}{\emph{Sociological perspectives}} \bibinfo{volume}{60}, \bibinfo{number}{4} (\bibinfo{year}{2017}), \bibinfo{pages}{810--833}.
\newblock


\bibitem[Surden(2021)]%
        {surden2021machine}
\bibfield{author}{\bibinfo{person}{Harry Surden}.} \bibinfo{year}{2021}\natexlab{}.
\newblock \showarticletitle{Machine learning and law: An overview}.
\newblock \bibinfo{journal}{\emph{Research handbook on big data law}} (\bibinfo{year}{2021}), \bibinfo{pages}{171--184}.
\newblock


\bibitem[Team et~al\mbox{.}(2024)]%
        {team2024gemini}
\bibfield{author}{\bibinfo{person}{Gemini Team}, \bibinfo{person}{Petko Georgiev}, \bibinfo{person}{Ving~Ian Lei}, \bibinfo{person}{Ryan Burnell}, \bibinfo{person}{Libin Bai}, \bibinfo{person}{Anmol Gulati}, \bibinfo{person}{Garrett Tanzer}, \bibinfo{person}{Damien Vincent}, \bibinfo{person}{Zhufeng Pan}, \bibinfo{person}{Shibo Wang}, {et~al\mbox{.}}} \bibinfo{year}{2024}\natexlab{}.
\newblock \showarticletitle{Gemini 1.5: Unlocking multimodal understanding across millions of tokens of context}.
\newblock \bibinfo{journal}{\emph{arXiv preprint arXiv:2403.05530}} (\bibinfo{year}{2024}).
\newblock


\bibitem[Viswanath and Zhang(2025)]%
        {viswanath2025fairpytoolkitevaluationprediction}
\bibfield{author}{\bibinfo{person}{Hrishikesh Viswanath} {and} \bibinfo{person}{Tianyi Zhang}.} \bibinfo{year}{2025}\natexlab{}.
\newblock \bibinfo{title}{FairPy: A Toolkit for Evaluation of Prediction Biases and their Mitigation in Large Language Models}.
\newblock
\showeprint[arxiv]{2302.05508}~[cs.CL]
\urldef\tempurl%
\url{https://arxiv.org/abs/2302.05508}
\showURL{%
\tempurl}


\bibitem[{Vite Contributors}(2020)]%
        {vite2020}
\bibfield{author}{\bibinfo{person}{{Vite Contributors}}.} \bibinfo{year}{2020}\natexlab{}.
\newblock \bibinfo{title}{Vite -- Next Generation Frontend Tooling}.
\newblock \bibinfo{howpublished}{\url{https://github.com/vitejs/vite}}.
\newblock
\newblock
\shownote{Accessed: 16 May 2025}.


\bibitem[{Vue.js Developers}(2014)]%
        {vue3}
\bibfield{author}{\bibinfo{person}{{Vue.js Developers}}.} \bibinfo{year}{2014}\natexlab{}.
\newblock \bibinfo{title}{Vue.js -- The Progressive JavaScript Framework v3.0}.
\newblock \bibinfo{howpublished}{\url{https://vuejs.org/guide/introduction.html}}.
\newblock
\newblock
\shownote{Accessed: 16 May 2025}.


\bibitem[Washington(2018)]%
        {washington2018argue}
\bibfield{author}{\bibinfo{person}{Anne~L Washington}.} \bibinfo{year}{2018}\natexlab{}.
\newblock \showarticletitle{How to argue with an algorithm: Lessons from the COMPAS-ProPublica debate}.
\newblock \bibinfo{journal}{\emph{Colo. Tech. LJ}}  \bibinfo{volume}{17} (\bibinfo{year}{2018}), \bibinfo{pages}{131}.
\newblock


\bibitem[Weerts et~al\mbox{.}(2023)]%
        {Weerts_Fairlearn_Assessing_and_2023}
\bibfield{author}{\bibinfo{person}{Hilde Weerts}, \bibinfo{person}{Miroslav Dudík}, \bibinfo{person}{Richard Edgar}, \bibinfo{person}{Adrin Jalali}, \bibinfo{person}{Roman Lutz}, {and} \bibinfo{person}{Michael Madaio}.} \bibinfo{year}{2023}\natexlab{}.
\newblock \showarticletitle{{Fairlearn: Assessing and Improving Fairness of AI Systems}}.
\newblock \bibinfo{journal}{\emph{Journal of Machine Learning Research}}  \bibinfo{volume}{24} (\bibinfo{year}{2023}).
\newblock
\urldef\tempurl%
\url{http://jmlr.org/papers/v24/23-0389.html}
\showURL{%
\tempurl}


\bibitem[Wilson and Caliskan(2024)]%
        {wilson2024gender}
\bibfield{author}{\bibinfo{person}{Kyra Wilson} {and} \bibinfo{person}{Aylin Caliskan}.} \bibinfo{year}{2024}\natexlab{}.
\newblock \showarticletitle{Gender, race, and intersectional bias in resume screening via language model retrieval}. In \bibinfo{booktitle}{\emph{Proceedings of the AAAI/ACM Conference on AI, Ethics, and Society}}, Vol.~\bibinfo{volume}{7}. \bibinfo{pages}{1578--1590}.
\newblock


\bibitem[Xu(2021)]%
        {xu2021pystata}
\bibfield{author}{\bibinfo{person}{Zhao Xu}.} \bibinfo{year}{2021}\natexlab{}.
\newblock \showarticletitle{PyStata-Python and Stata integration}. In \bibinfo{booktitle}{\emph{London Stata Conference 2021}}. Stata Users Group.
\newblock


\bibitem[Xue et~al\mbox{.}(2024)]%
        {xue2024leec}
\bibfield{author}{\bibinfo{person}{Zongyue Xue}, \bibinfo{person}{Huanghai Liu}, \bibinfo{person}{Yiran Hu}, \bibinfo{person}{Yuliang Qian}, \bibinfo{person}{Yajing Wang}, \bibinfo{person}{Kangle Kong}, \bibinfo{person}{Chenlu Wang}, \bibinfo{person}{Yun Liu}, {and} \bibinfo{person}{Weixing Shen}.} \bibinfo{year}{2024}\natexlab{}.
\newblock \showarticletitle{LEEC for Judicial Fairness: A Legal Element Extraction Dataset with Extensive Extra-Legal Labels}. In \bibinfo{booktitle}{\emph{Proceedings of the Thirty-Third International Joint Conference on Artificial Intelligence, IJCAI-24}}. \bibinfo{pages}{7527--7535}.
\newblock


\bibitem[Yang et~al\mbox{.}(2024)]%
        {yang2024qwen25mathtechnicalreportmathematical}
\bibfield{author}{\bibinfo{person}{An Yang}, \bibinfo{person}{Beichen Zhang}, \bibinfo{person}{Binyuan Hui}, \bibinfo{person}{Bofei Gao}, \bibinfo{person}{Bowen Yu}, \bibinfo{person}{Chengpeng Li}, \bibinfo{person}{Dayiheng Liu}, \bibinfo{person}{Jianhong Tu}, \bibinfo{person}{Jingren Zhou}, \bibinfo{person}{Junyang Lin}, \bibinfo{person}{Keming Lu}, \bibinfo{person}{Mingfeng Xue}, \bibinfo{person}{Runji Lin}, \bibinfo{person}{Tianyu Liu}, \bibinfo{person}{Xingzhang Ren}, {and} \bibinfo{person}{Zhenru Zhang}.} \bibinfo{year}{2024}\natexlab{}.
\newblock \bibinfo{title}{Qwen2.5-Math Technical Report: Toward Mathematical Expert Model via Self-Improvement}.
\newblock
\showeprint[arxiv]{2409.12122}~[cs.CL]
\urldef\tempurl%
\url{https://arxiv.org/abs/2409.12122}
\showURL{%
\tempurl}


\end{thebibliography}

\end{document}